\documentclass{dhbenelux}

\usepackage{amsmath}
\usepackage{array} 
\usepackage{longtable}
 
\usepackage{booktabs} 
\usepackage{authblk} 
\usepackage{graphicx} 

\author[,a,b]{Chao Huang\textsuperscript{1,2}}
\author[,b]{Huichen Xiao\textsuperscript{1}}
\author[b]{Chen Chen}
\author[b]{Chunyan Chen}
\author[b]{Yi Zhao}
\author[c,d]{Shiyu Du}
\author[c]{Yiming Zhang}
\author[b]{He Sha}
\author[b]{Ruixin Gu}

\affil[a]{Institute of Computing Technology, Chinese Academy of Science, Beijing, China}
\affil[b]{Ningbo Institute of Information Technology Application, Chinese Academy of Sciences (CAS), Ningbo, China}
\affil[c]{Engineering Laboratory of Advanced Energy Materials Ningbo Institute of Materials Technology and Engineering Chinese Academy of Sciences, Ningbo, China}
\affil[d]{School of Materials Science and Engineering and School of Computer Science, China University of Petroleum (East China), Qingdao, China}

\title{Polymetis:Large Language Modeling for Multiple Material Domains}

\begin{document}
\maketitle
\footnotetext[1]{These authors have made equal contributions}
\footnotetext[2]{Corresponding author: chuang@ict.ac.cn}

\begin{abstract}
    \noindent {As the application of large language models in various fields continues to expand, materials science also ushers in opportunities for AI-driven innovation. The traditional way of relying on manual search for materials science-related information is now using artificial intelligence technology as an auxiliary tool to improve the efficiency of materials science research. To accelerate researchers' knowledge acquisition and intelligent decision-making support in materials science research, this paper proposes a large language model Polymetis model for a variety of materials fields, aiming to provide highly professional knowledge answers in the field of materials, covering energy materials, functional materials, alloy materials, physical chemistry, biology, and other material directions. The model uses a dataset of about 2 million material knowledge instructions, and in the process of building the dataset, we developed the Intelligent Extraction Large Model (IELM), which is specially used to extract and form structured knowledge from scientific texts, avoiding a large number of costs that need to be manually annotated, and improving efficiency. We inject this data into the GLM4-9B model for learning to enhance its inference capabilities in a variety of material domains. In addition, we have introduced enhanced prompt strategies to ensure that the answers to the model are more organized and comprehensive, providing efficient and comprehensive intelligent support for the diverse needs of materials science exploration, and promoting the development of material science.}
\end{abstract}

\section{Introduction}

Large-scale language models (LLMs) have laid a solid foundation for various applications. ChatGPT and GPT-4.0 \citet{achiam2023gpt}, developed by OpenAI, have 175 billion and 18 trillion parameters, respectively, and have a wide range of applications in natural language processing. And the details of their training methods are not publicized. The GLM base model  from Tsinghua University provides a compelling option for natural language processing\citet{du2021glm, ouyang2022training}. It supports both English and Chinese, with high accuracy, cross-platform compatibility, repeatability, and fast inference. Ernie 3.0Titan, introduced by Baidu, is an upgraded version of the Ernie family of models \citet{sun2019ernie,sun2020ernie,sun2021ernie}, which employs deeper knowledge fusion techniques with tens of billions of references to support multiple language comprehension and generation tasks. The emergence of open-source alternative models, such as LLaMA \citet{touvron2023llama} and RWKV\citet{peng2023rwkv}, provides a variety of options for fine-tuning the underlying large-scale language models (e.g., Alpaca \citet{taori2023stanford} and Vicuna\citet{chiang2023vicuna}). However, most of the instruction datasets are self-generated by GPT-4, which may lead to reduced model inference accuracy \citet{wang2022self,jablonka2023gpt}. The traditional manual extraction methods are inefficient and costly, which poses a great challenge for material science text mining \citet{kononova2021opportunities}.

In materials-specific domains, LLM has shown great potential in materials science analytics, bringing significant innovations to scientific research and industrial applications. Models such as DARWIN SERIES \citet{xie2023darwin} and MatChat \citet{chen2023matchat} have demonstrated the potential of AI to assist in the materials domain, providing researchers with accelerated access to domain-specific information \cite{weston2019named} and providing knowledge extraction and discovery processes \citet{venugopal2021looking,fang2023mol}. However, it has some minor drawbacks. The DARWIN model suffers from catastrophic forgetting problems, the generalization ability is greatly reduced, and the answers lack organization and precision, in addition, DARWIN uses multiple models instead of a unified model which may lead to complexity and potential inefficiency. As for MatChat, its first limitation for Chinese users is that it only supports English usage and response accuracy is a tricky issue when dealing with unorganized datasets.

Although these models have made important contributions to the field of materials science, they generally suffer from several limitations: first, catastrophic forgetting problems and lack of generalization capabilities. Second, the downstream tasks performed by these models are more limited and cannot accommodate the multi-domain materials knowledge dialog. These issues make existing models often fail to meet the needs of researchers when dealing with more complex and diverse material knowledge queries.

To overcome these challenges, we propose Polymetis, a multidisciplinary materials science-oriented large language model designed to provide specialized knowledge support covering multiple directions such as energy materials, functional materials, alloy materials, physical chemistry, biomaterials, etc. Polymetis uses a dataset of about 2 million knowledge instructions in the materials domain and utilizes our self-developed Intellectual Extractive Large Model (IELM) model is used to automatically extract and process structured material domain knowledge from material science literature, avoiding the need for large amounts of manual annotation and improving the efficiency of dataset construction.
The advantages of Polymetis are:

a) First, we developed the IELM model, through which automated knowledge extraction and data processing techniques enable the model's training dataset to be rapidly extended to multiple material domains, improving the broad applicability of the Polymetis model;

b) We utilize the GLM4-9B model for parameter-efficient Lora fine-tuning and introduce an enhanced prompt strategy in order to improve the organization and accuracy of the trained model's inference results. Using the benchmark answers provided by materials science experts for comparative evaluation, Polymetis demonstrates excellent complex instruction understanding and multi-domain reasoning capabilities, which can provide researchers with accurate and efficient materials knowledge exploration for their needs.

\section{Methods}

In this section, we will introduce the process of constructing the instruction dataset, which includes the creation of question and answer pairs, the development of the Intellectual Extractive Large Model IELM, and the construction of the material knowledge instruction dataset. Additionally, we will describe the training process of the material knowledge Big Model Polymetis.
The following is a general technical flowchart of the method, as shown in Figure~\ref{fig:model_maxwell}:

\begin{figure}[ht]
\begin{center}
\includegraphics[width=1\linewidth]{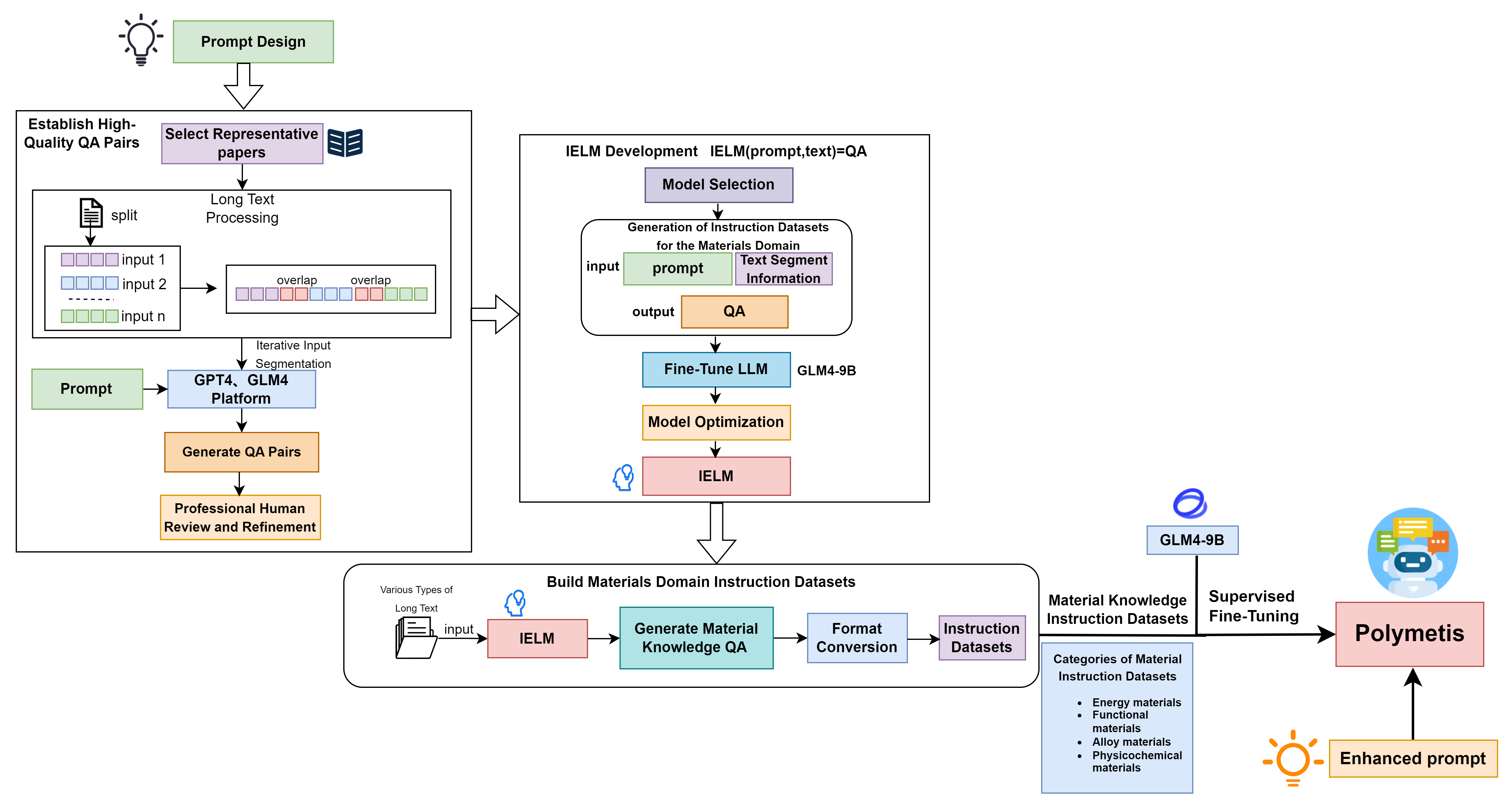}
\end{center}
\caption{Polymetis model development process.}
\label{fig:model_maxwell}
\end{figure}

\subsection{Building Instruction Datasets}

\subsubsection{Establishment of QA pairs}

Select representative and widely used text from scientific research literature in multiple fields, covering such related fields as energy materials, functional materials and so on. These texts are processed and partitioned into multiple text segments. Using the GPT4 open platform API interface, iteratively input these text segments into the GPT4 model. During the input process, structured prompts are incorporated to guide the GPT4 model in generating QA pairs associated with each text segment, structured to include role responsibilities, requirements, and output formatting standards. The generated material knowledge QA pairs are reviewed and refined by specialized material science professionals to form a high-quality collection of QA pairs.

\subsubsection{Development IELM}

Developing the IELM of the Intelligent Extraction Large Model: 
\begin{itemize}
    \item \textbf{a) Constructing the instruction generation dataset:} Based on the generated high-quality Q\&A pairs, structured prompts, and their corresponding text segments, a high-quality instruction generation dataset of the material domain is constructed,with
    \[
    \text{IELM} = (\text{prompt}, \text{text}) = \text{QA},
    \]
    where \text{prompt} and \text{text} are the instruction requirements and text segments issued to the model, respectively, and \text{QA} is the corresponding QA generated by the model.
    
    \item \textbf{b) Select the base model for IELM:} Select an appropriate model as the base model for training by comparing the Inference Performance of multiple open-source models. Finally, GLM4-9B is chosen as the base model for training IELM.

    \item \textbf{c) IELM model:} The base model is trained using the constructed high-quality domain knowledge instruction generation dataset. The large model training hyperparameters are continuously adjusted through multiple rounds of testing and based on the model performance, thus improving the performance of the model in the QA generation task. Eventually, we developed an intelligent large model specialized for Q\&A generation, called IELM (Intelligent Extraction Large Model). The command adherence ability of this model is greatly improved compared to the charm modeling platform, and it is able to strictly follow the output format of the prompts for QA generation, whereas the output of the ChatGLM modeling platform lacks some of the required symbols and details, and the QA pairs generated by the IELM model under the guidance of the structured prompts are much richer and more specific in terms of information.
\end{itemize}

\subsubsection{Constructing material knowledge instruction dataset}

Utilizing the trained IELM of the Wisdom Diaspora Big Model, efficient and automated QA generation is carried out on the textual data in the material domain. This process can automatically generate high-quality domain knowledge QA pairs for different types of materials domain literature, teaching materials, etc., which in turn provides a reliable data source for the training of the materials knowledge big model. At the same time, the generated QA pairs are formatted according to the predefined format requirements of specialized domain datasets to meet the specifications of the training dataset, and after format conversion, these QA pairs are integrated into a structured instruction dataset, which can be directly used for the training of the big model of material domain knowledge.

\subsection{Polymetis model development process}

\subsubsection{Base Model}

GLM4-9B is a new generation of the pre-trained model introduced by Zhipu AI, which belongs to the open source version in the GLM-4 series. The model performs well in several aspects.GLM-4-9B is trained using 10T of high-quality multilingual data, which is more than three times as much as the ChatGLM3-6B model, and employs FP8 technology for efficient pre-training.The model has about 9.4B (9.4 billion) parameters. In this paper, we will use the material knowledge instruction dataset constructed in the previous section to fine-tune the GLM4-9B model based on the GLM4-9B model in order to make it perform better in the field of material science.

\subsubsection{Structured material datasets}

Our data is a structured instruction dataset constructed from about 2 million QA pairs obtained by extracting QA from about 100,000 papers and processing and integrating them using the IELM Intelligent Extractive Large Model developed in Section 2.1. It contains knowledge of about 10 material domains, including energy materials, functional materials, alloy materials, nanomaterials, biomaterials, applied polymer materials, chemical-physical materials, etc., and the distribution of the data is shown in Figure~\ref{fig:data_distri} below. The format of the instruction dataset used for training is as follows, and the specific examples are shown in Appendix~\ref{app:instruction_examples}:

{"messages": [{"role": "user", "content": ""}, {"role": "assistant", "content": ""}]}

\begin{figure}[ht]
\begin{center}
\includegraphics[width=1\linewidth]{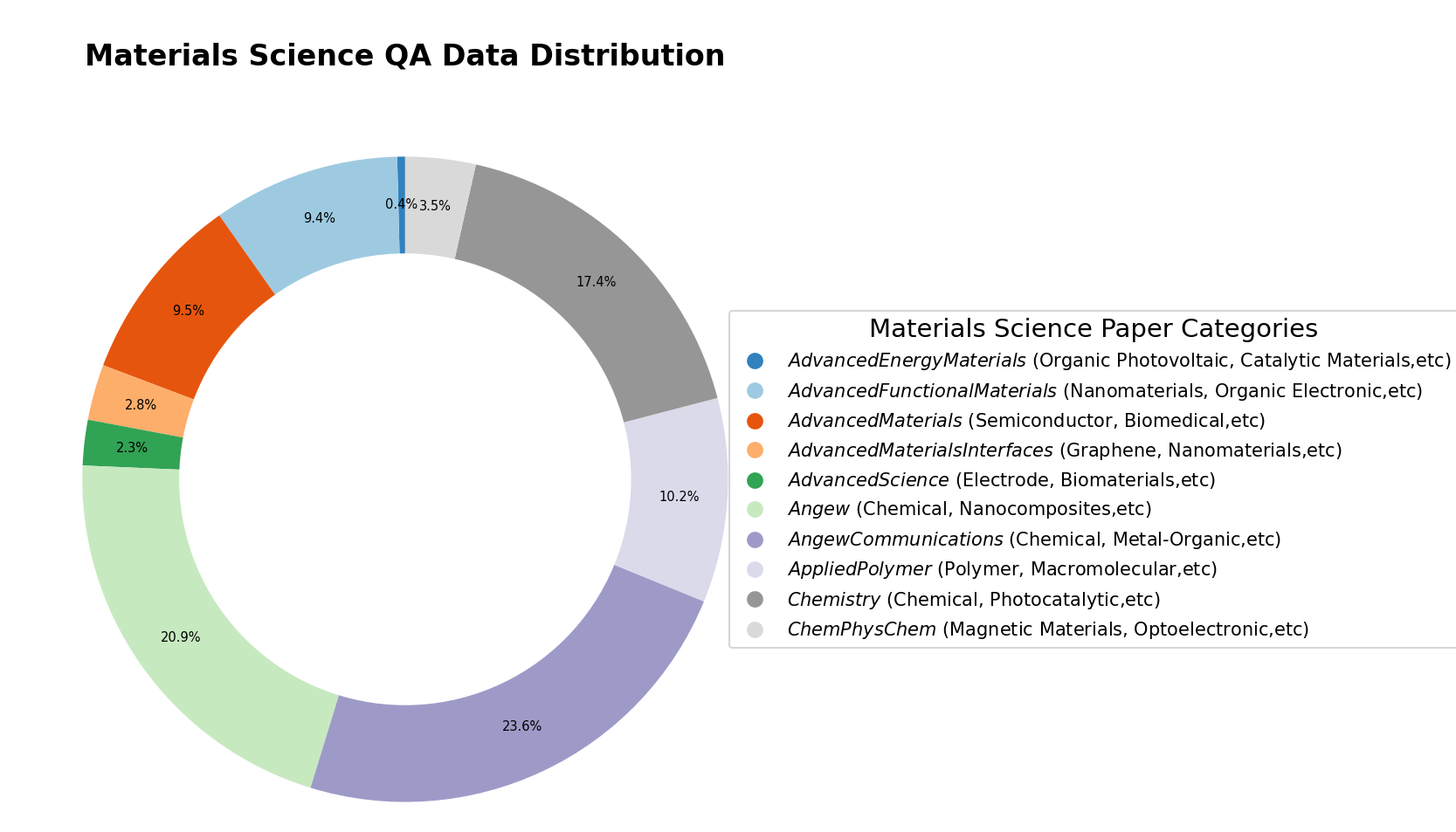}
\end{center}
\caption{QA data distribution.}
\label{fig:data_distri}
\end{figure}

\subsubsection{Training Process}

In our study, we fine-tuned the GLM4-9B model based on the LlamaFactory framework using four NPUs on a Huawei Ascend 800T A2 training server. To prevent the catastrophic forgetting problem, we used the parameter-efficient fine-tuning method LoRA (Low-Rank Adaptation). This method achieves fine-tuning by freezing most of the parameters of the model and adding only a small number of learnable low-rank matrices. This approach avoids updating the core weights of the model, greatly reduces the number of parameters that need to be trained and provides more efficient and economical fine-tuning.

In the fine-tuning process, we use the following key hyperparameters: the learning rate is set to 1e-5, the batch size is 4, and 3 epochs are trained. The fine-tuned model, named Polymetis, provides a more efficient adaptation to a variety of materials knowledge-questioning tasks (e.g., energy, functional, alloying materials, etc.), while preserving the knowledge structure of the original model, achieving a balance between accuracy and economy.

\subsubsection{Enhanced Prompt}

After fine-tuning supervised instructions, many large language models (LLMs) may have some common problems when outputting content, such as incoherent responses, loose structure, or imprecise expressions. In order to overcome these problems and improve the performance of the model in a variety of material knowledge question-answering tasks, we introduced an enhanced prompt strategy based on the trained model to optimize the output quality of the trained model.

Specifically, we carefully designed and tuned system prompts to guide the model to better understand the task requirements and provide clearer and more structured answers based on contextual information and mission objectives. First, we clarified the mission objectives and requirements to ensure that the model could understand the context and requirements of the problem. Next, by setting the role of the model, we provide a framework for it so that its output is tailored to the needs of a particular domain. We set up the model as a material science expert so that the response style of the model is more appropriate to the terminology and body of knowledge in the field. Through this role-setting, the trained model can not only avoid generating overly generic answers but also more accurately capture the key concepts and terms in the domain, thereby improving the professionalism and pertinence of the answers.

In addition, to ensure that the output content of the model is organized, we clarified the logical structure of the model's answers and required the model to follow certain steps when answering, such as expanding the specific details in detail first and then giving a concise conclusion, to make the final answer more efficient and grasp the key points. Boundary conditions have also been added, such as avoiding vague answers or verbose extraneous details. This enhanced prompt setting helps the model to present information in a more organized and professional manner, thereby effectively improving the logic and practicality of the answer.

\section{Experiments}

\subsection{Baselines}

Currently, there is a lack of large models specifically designed for multi-material knowledge question-answering tasks. In this experiment, we compare our Polymetis large model with other mainstream general-purpose models for multi-material knowledge-based question answering.

\begin{itemize}

\item[$\bullet$] \textbf{ChatGPT-3.5:}Developed by OpenAI, based on the GPT-3.5 (Generative Pre-trained Transformer 3.5) architecture, with approximately 175 billion parameters. This model is capable of complex conversational reasoning, text generation, and understanding.
\item[$\bullet$] \textbf{Ernie Bot:} Developed by Baidu, based on the ERNIE (Enhanced Representation through Knowledge Integration) large model architecture. The ERNIE series integrates deeper knowledge fusion techniques, with parameters reaching hundreds of billions. It supports a wide range of language understanding and generation tasks, excelling particularly in natural language generation and question answering in Chinese contexts.
\item[$\bullet$] \textbf{ChatGLM:}Developed by Zhihu AI, this is a Chinese large language model based on their self-developed GLM (General Language Model) architecture, with over 100 billion parameters. The model specializes in multi-task learning, knowledge reasoning, and text generation in Chinese.
\item[$\bullet$] \textbf{Qwen:}Developed by Alibaba, this model has a parameter scale in the hundred-billion range (specific parameter details are not disclosed).
\end{itemize}

\subsection{Metrics}

In the process of text processing and model evaluation, it is an important task to evaluate the proximity between the model output and the actual answer. Especially in fields such as materials science, benchmark answers provided by experts are often used to measure the performance of models. However, due to the complexity and diversity of natural language, even two seemingly identical answers may differ in the way they are expressed and worded, but their actual semantics are the same. Therefore, traditional methods based on exact matching may not be able to effectively capture this semantic similarity, which will affect the evaluation results of the model.

To ensure the objectivity and accuracy of the assessment, our benchmark answers are obtained through a series of carefully designed processes. Specifically, the expert first conducts an in-depth data review of each question and combines the outputs of multiple AI tools to synthesize the answer that best matches the domain knowledge. In this process, the expert does not know which model corresponds to each output result, thus avoiding the influence of human bias. Based on the reasonableness and accuracy of each model's output, the expert appropriately adopts the answers that are in line with real material science knowledge and finally forms a comprehensive benchmark answer. This approach ensures the professionalism and objectivity of the benchmark answer and helps to improve the accuracy of the assessment by referring to multiple perspectives.

To address the limitations of traditional exact matching methods, the use of semantic similarity to measure the proximity between the model output and the benchmark answers is a more appropriate choice.\citet{wang2019evaluating}. Semantic similarity not only considers superficial lexical matching, but also identifies the semantic similarity of different words and phrases, which is essential for dealing with synonyms, grammatical variations, and contextual relationships in natural language. In this study, we chose to evaluate the performance of the Polymetis model against other models by calculating the similarity of semantic vectors. With this approach, we are able to measure the quality of the model's output more accurately, especially in the face of linguistic differences but semantic similarities, avoiding evaluation bias due to superficial lexical inconsistencies.

In addition, as a core tool in natural language processing (NLP), the word vector model can convert text in natural language into word vector representations in semantic space, so as to complete various text information processing tasks in vector space. From word2vec\citet{church2017word2vec}, and ELMo, to the proposal and development of BERT, the word vector model marks an important breakthrough in the field of natural language processing. In particular, the Bidirectional Encoder Representations from Transformers (BERT) language model proposed by Devlin et al. in 2018 has performed well in multiple tasks of natural language processing\citet{devlin2018bert}, especially in typical applications such as text classification, sentiment analysis, and question answering systems. BERT is trained by a bidirectional encoder to capture contextual information more effectively, making it perform well in multiple downstream tasks. In these tasks, the eigenvectors of BERT can be directly embedded as words of the task, so as to provide a more accurate representation of the specific task.
 
When computing semantic similarity, a commonly used method in natural language processing is Cosine Similarity, also known as Cosine Distance. Cosine similarity evaluates the degree of similarity between two vectors by measuring their directional consistency, focusing on the relative differences between dimensions in the vector space rather than numerical differences. Specifically, the formula for cosine similarity is as follows:

\[
\text{cosine}(x, y) = \frac{x \cdot y}{\|x\| \|y\|}
\]

where \( x \cdot y = \sum_{i} x_i y_i \) represents the dot product of the vectors \(x\) and \(y\), and \( \|x\| = \sqrt{\sum_{i} x_i^2} \) and \( \|y\| = \sqrt{\sum_{i} y_i^2} \) represent the Euclidean norms of the vectors \(x\) and \(y\), respectively. The closer the calculated cosine value is to 1, the more similar the two vectors are. The closer the cosine value is to 0, the less similar the two vectors are. By using cosine similarity, we are able to effectively quantify the similarity between two-word vectors, thus providing a reliable metric for evaluating the performance of the Polymetis model against other models.

\section{Results}

In this section, the results of comparing the semantic similarity between the responses of our Polymetis model and other mainstream models for each question with the benchmark answers will be shown separately, as shown in Table~\ref{tab:comparison_results}, if the semantic similarity value of the model's answer and the benchmark answer is closer to 1, it means that it is closer to the benchmark answer, indicating that the responses it provides are relatively more accurate and comprehensive. The input questions, benchmark answers, and responses output by each model are shown in Appendix~\ref{app:model_outputs}:

\begin{table}[h]
\centering
\caption{Comparison of the answers' similarity between the Polymetis model and other models.}
\begin{tabular}{|l|c|c|c|}
\hline
\textbf{} & \textbf{Question 1} & \textbf{Question 2} & \textbf{Question 3} \\
\hline
\textbf{Benchmarke Answer}  & 1.0000 & 1.0000 & 1.0000 \\
\hline
\textbf{ChatGPT-3.5}      & 0.8688 & 0.9250 & 0.9165 \\
\hline
\textbf{Qwen}             & 0.8938 & 0.9296 & 0.8784 \\
\hline
\textbf{Ernie Bot}        & 0.8884 & 0.9102 & 0.8206 \\
\hline
\textbf{ChatGLM}          & 0.8978 & 0.9052 & 0.8998 \\
\hline
\textbf{Polymetis}         & \textbf{0.9157} & \textbf{0.9342} & \textbf{0.9254} \\
\hline
\end{tabular}
\label{tab:comparison_results}
\end{table}

From the above similarity results, it can be seen that the Polymetis model outperforms other existing models on multiple problems, especially when dealing with expertise in the field of materials science. The Polymetis model has higher accuracy and comprehensiveness and can provide answers more in line with real-world needs. In addition, Polymetis can adapt to multiple materials domains, especially in reasoning across a wide range of domains such as energy materials, functional materials, alloy materials, and physical chemistry. By using specially designed IELM models to build structured knowledge datasets, Polymetis avoids the high cost of manual annotation and ensures that the data is of high quality and structured, thus enhancing learning.

\section{Conclusion}

This study presents the Polymetis model, which aims to provide accurate, well-organized, and professional knowledge responses in the field of materials science. By fine-tuning LoRA on the GLM4-9B model and combining it with an enhanced prompt strategy, Polymetis significantly improves reasoning in materials science tasks. In the process, we developed the IELM model for automated knowledge extraction, avoiding the need for extensive manual labeling of data while ensuring the high quality and structure of the training data.

Evaluations against benchmark answers provided by experts show that Polymetis outperforms existing models in several materials science subfields. Its efficient and scalable solution not only provides researchers with accurate and domain-relevant answers but also advances materials science research and intelligent decision-making. In the future, we will further expand the model's knowledge base and enhance its reasoning capabilities in more specialized domains to meet increasingly complex research needs.

\section{Limitations}

1.Coverage of the dataset: Although we used a dataset of about 2 million materials science-related knowledge instructions and extracted a large amount of structured materials knowledge through the IELM model, the dataset is still limited and cannot cover all possible research areas and complex expertise in materials science. Due to the continuous development of scientific research, the rapid changes in emerging fields and cutting-edge issues make it possible that the existing dataset may not be updated in time to include the latest scientific results and technological advances. Therefore, Polymetis' multi-domain knowledge datasets still need to be continuously expanded and optimized to adapt to the changing research needs.

2.The challenge of multi-domain integrated reasoning ability: The Polymetis model well demonstrates the reasoning capability in multiple materials science sub-domains; however, despite the model's better reasoning results in multiple domains, cross-domain reasoning tasks are still challenging. For example, certain interdisciplinary research problems may involve the integration and integrated reasoning of knowledge from multiple disciplines (e.g., materials physics, chemistry, biology, etc.), and the current model's reasoning capability may have some limitations in these complex cross-domain tasks.

\section{Acknowledgements}

This research was supported by the Engineering Laboratory of Advanced Energy Materials at the Ningbo Institute of Materials Technology and Engineering, the Chinese Academy of Sciences, the School of Materials Science and Engineering, and the School of Computer Science at the China University of Petroleum (East China). Their invaluable guidance and support have been instrumental in the success of this research.

\bibliography{references}

\appendix  

\section{Examples of Instruction Dataset}
\label{app:instruction_examples}
\begin{figure}[ht]
\begin{center}
\includegraphics[width=1\linewidth]{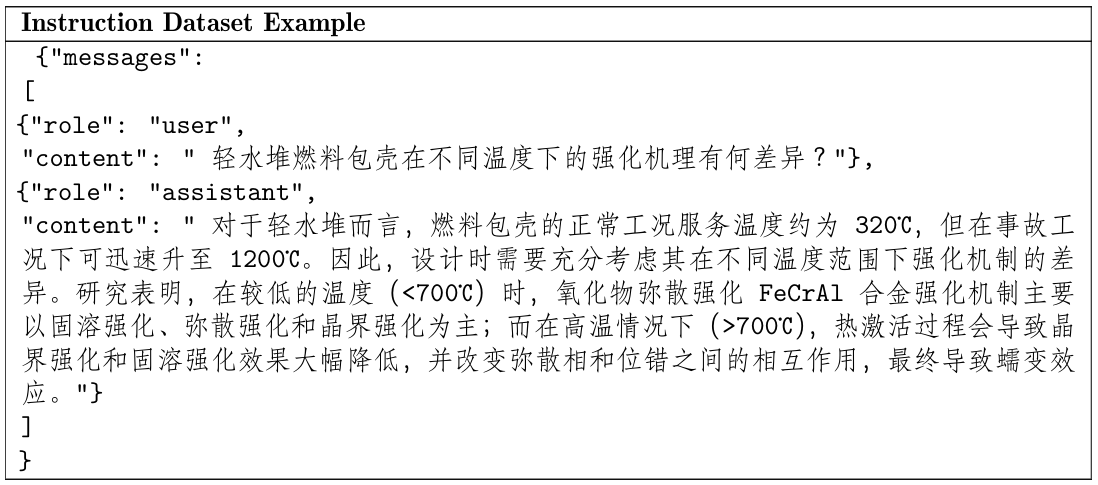}
\end{center}
\label{fig:instru_data}
\end{figure}

\section{Comparison of Model Outputs}
\label{app:model_outputs}
\begin{figure}[ht]
\begin{center}
\includegraphics[width=1\linewidth]{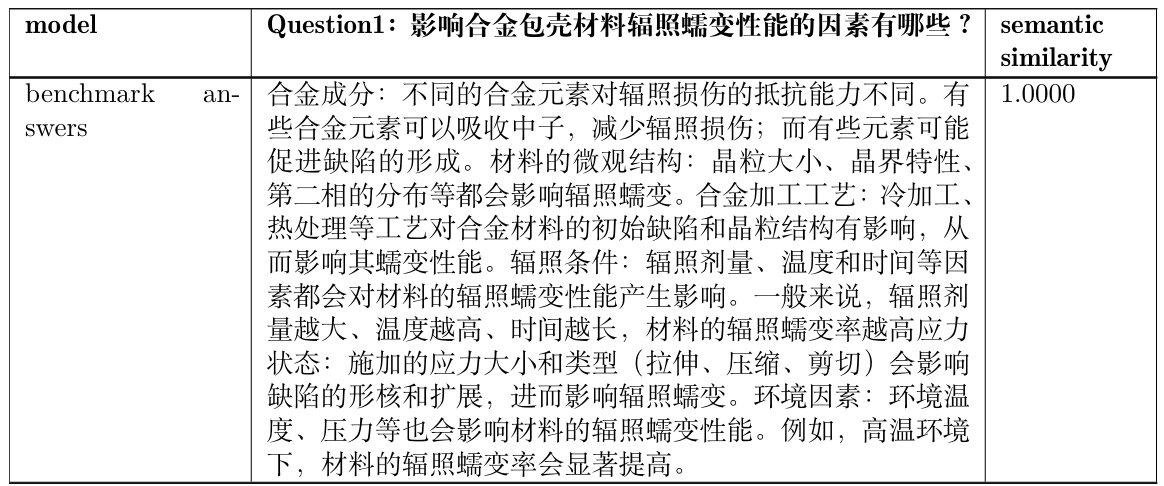}
\end{center}
\label{fig:a1}
\end{figure}

\begin{figure}[ht]
\begin{center}
\includegraphics[width=1\linewidth]{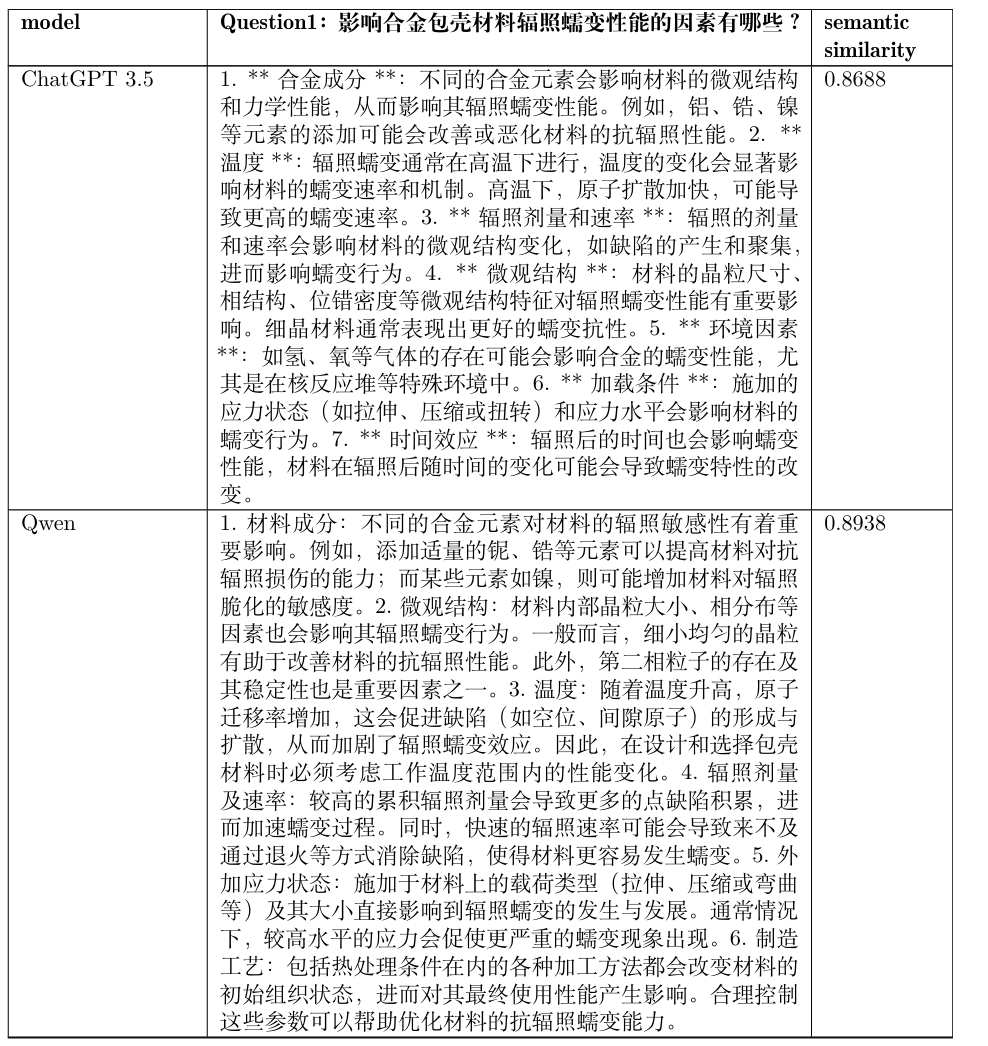}
\end{center}
\label{fig:a2}
\end{figure}

\begin{figure}[ht]
\begin{center}
\includegraphics[width=1\linewidth]{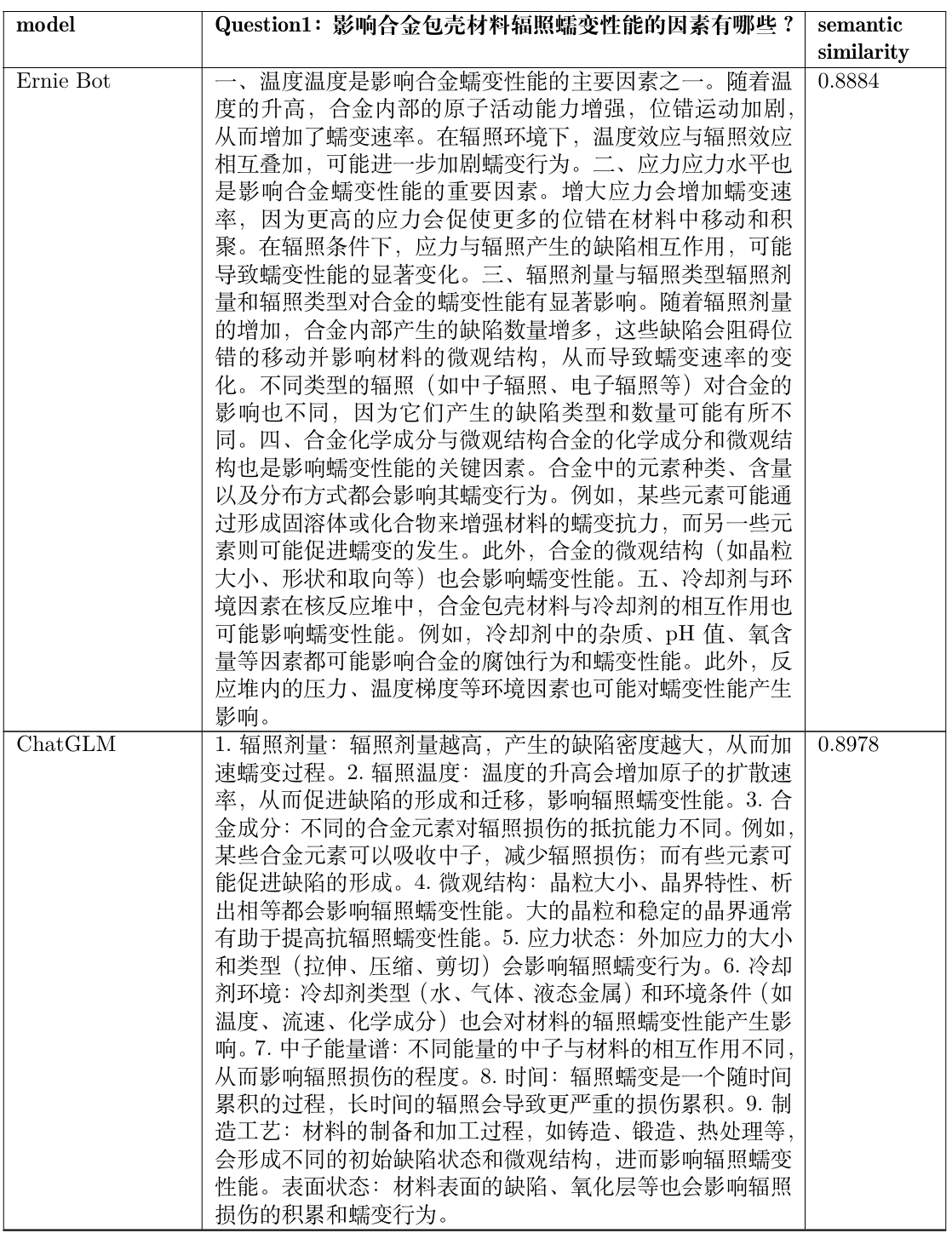}
\end{center}
\label{fig:a3}
\end{figure}

\begin{figure}[ht]
\begin{center}
\includegraphics[width=1\linewidth]{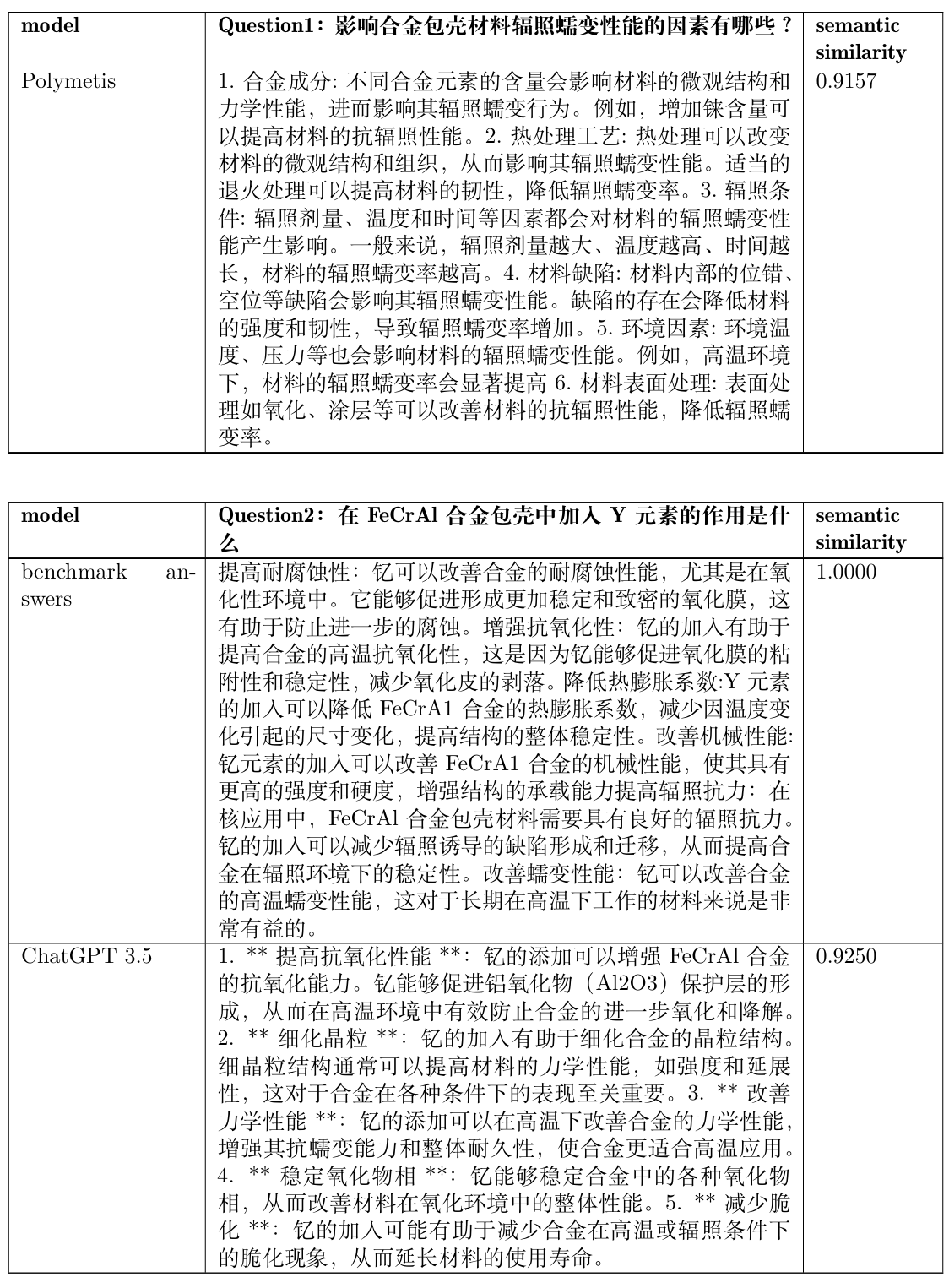}
\end{center}
\label{fig:a4}
\end{figure}

\begin{figure}[ht]
\begin{center}
\includegraphics[width=1\linewidth]{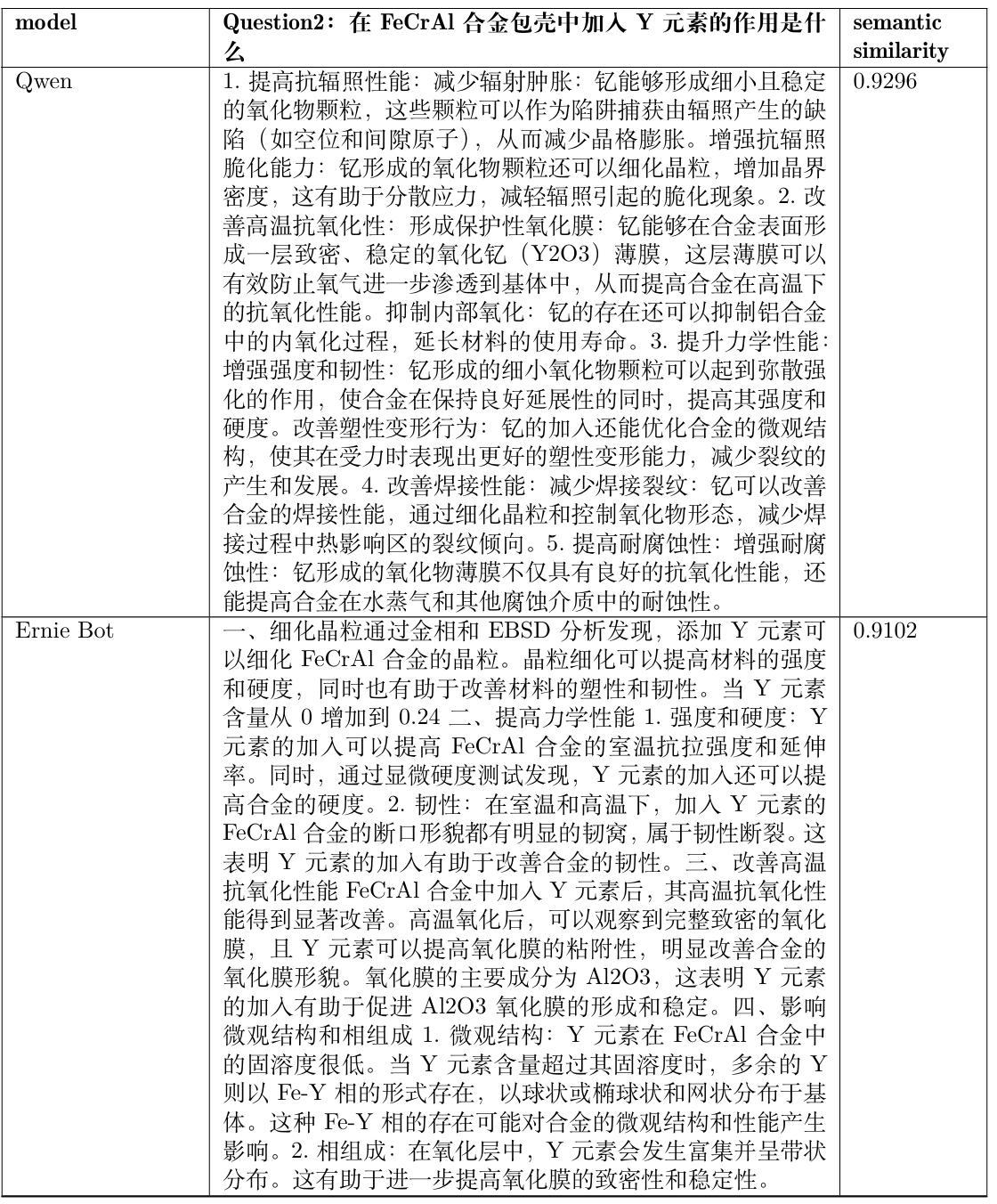}
\end{center}
\label{fig:a5}
\end{figure}

\begin{figure}[ht]
\begin{center}
\includegraphics[width=1\linewidth]{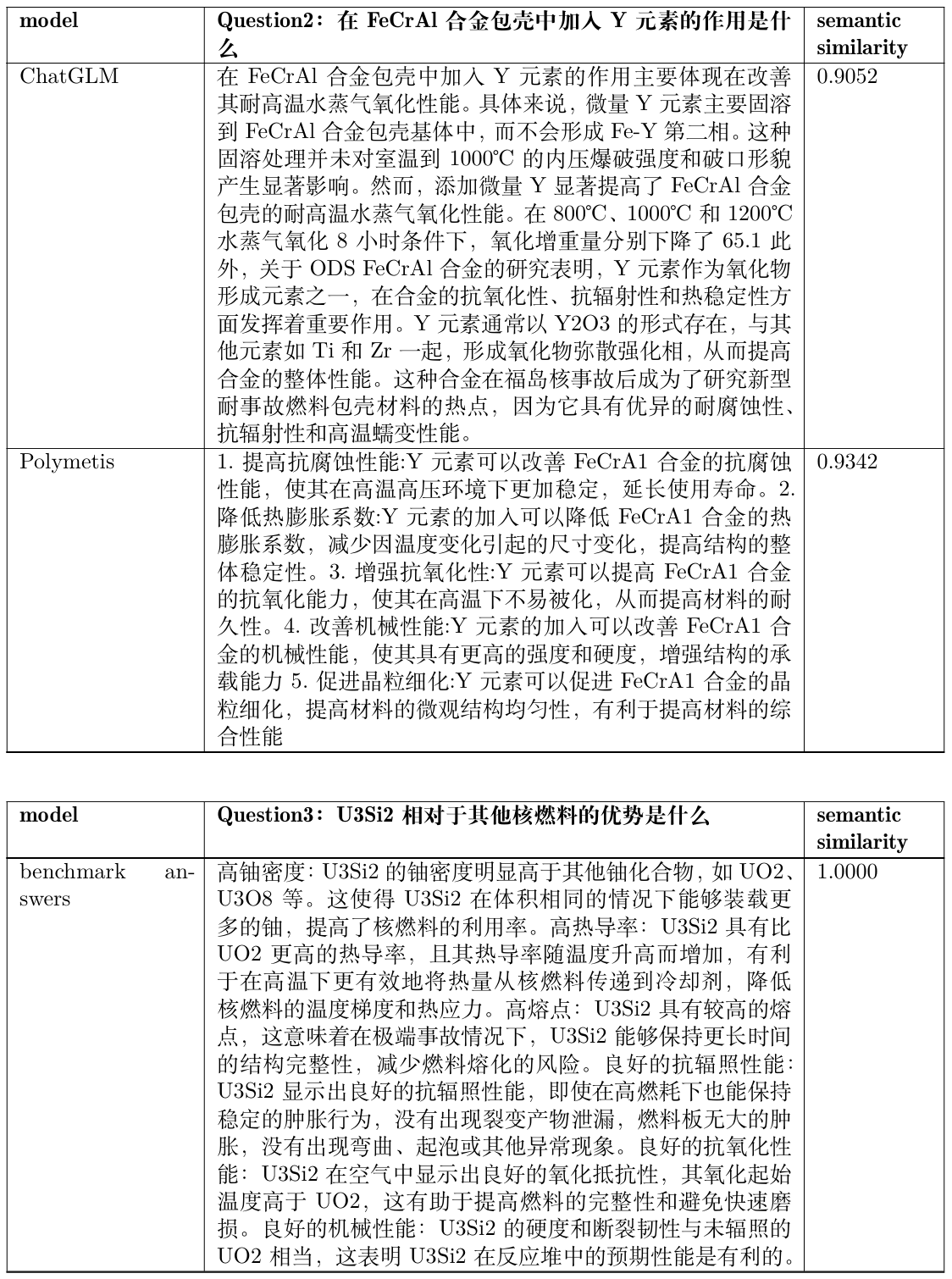}
\end{center}
\label{fig:a6}
\end{figure}

\begin{figure}[ht]
\begin{center}
\includegraphics[width=1\linewidth]{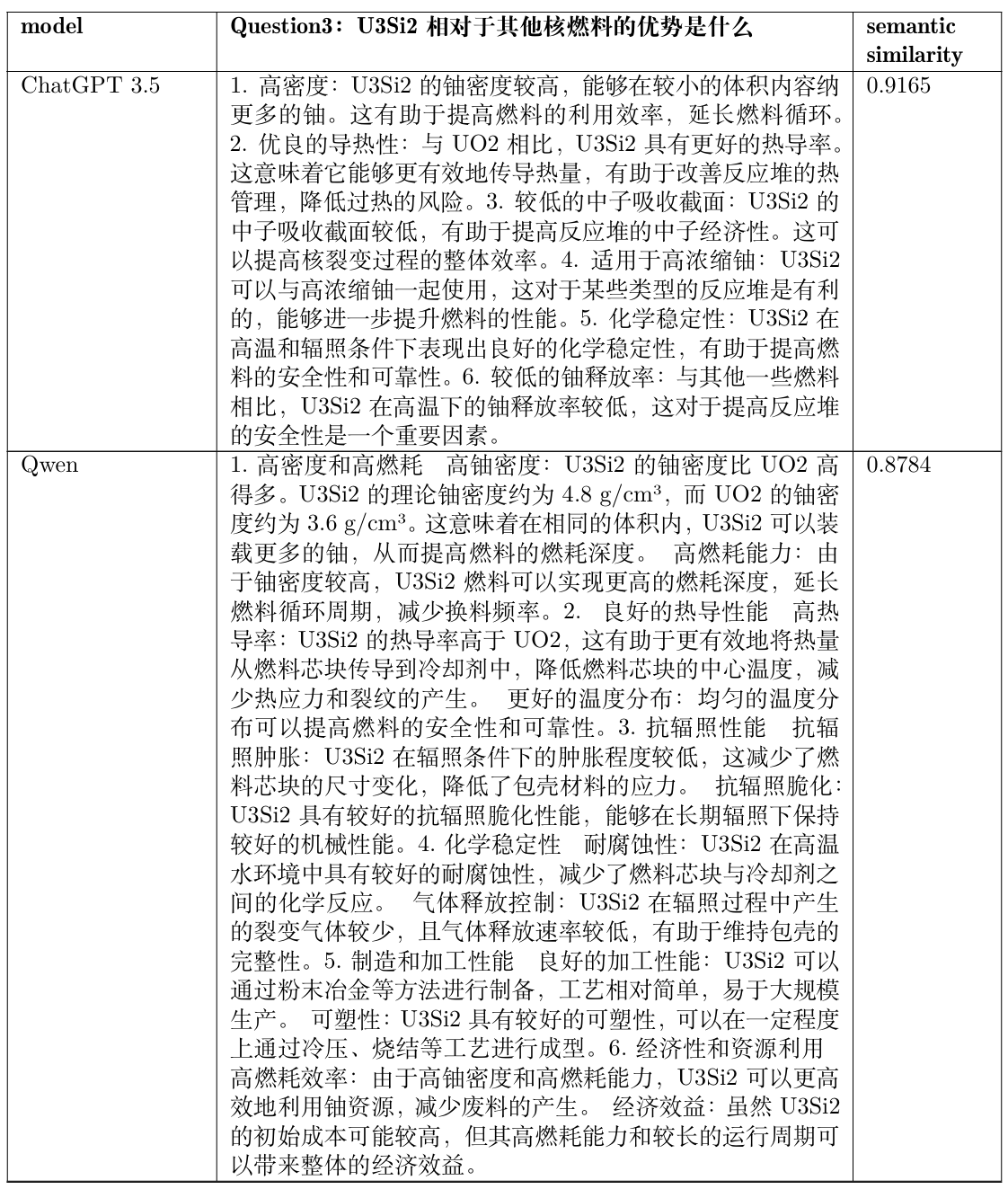}
\end{center}
\label{fig:a7}
\end{figure}

\begin{figure}[ht]
\begin{center}
\includegraphics[width=1\linewidth]{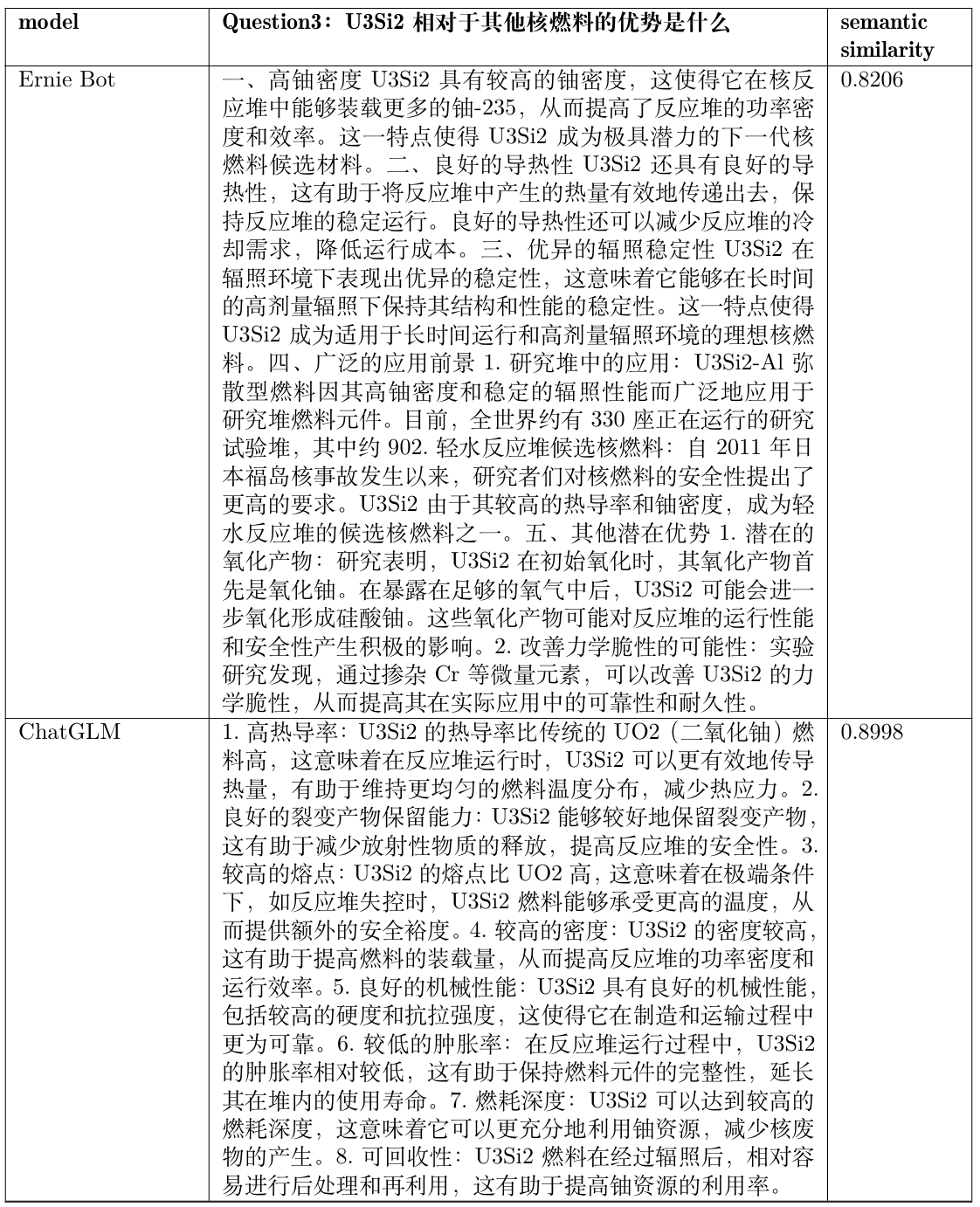}
\end{center}
\label{fig:a8}
\end{figure}

\begin{figure}[ht]
\begin{center}
\includegraphics[width=1\linewidth]{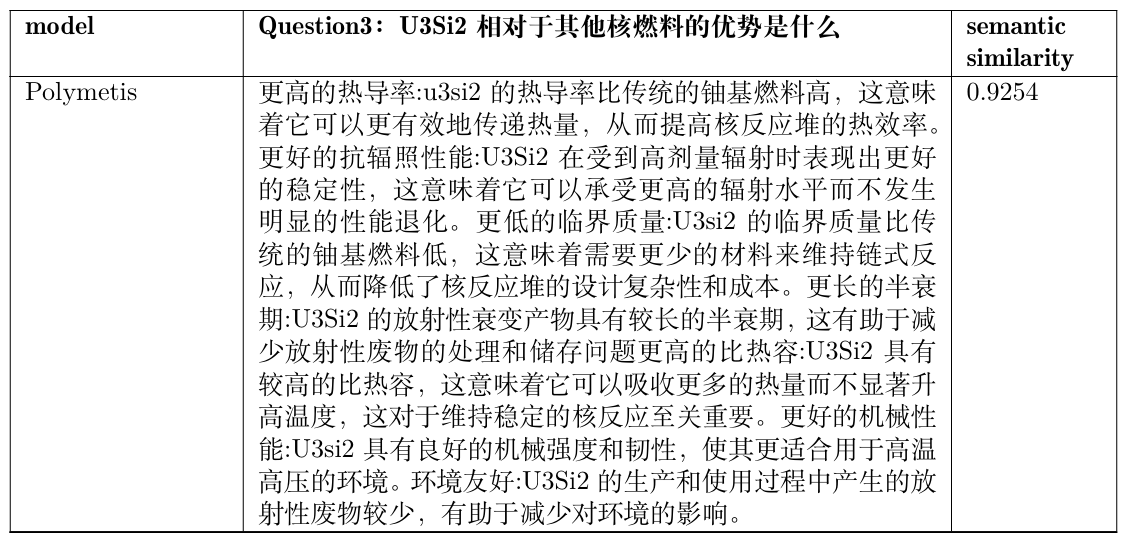}
\end{center}
\label{fig:a9}
\end{figure}

\end{document}